\documentclass[accepted]{uai2024} %

\usepackage{amsmath,amsfonts,amssymb}

\usepackage[american]{babel}

\usepackage{natbib} %
\renewcommand{\bibsection}{\subsubsection*{References}}
\usepackage{mathtools} %
\usepackage{booktabs} %
\usepackage{tikz} %
\usepackage{graphicx}
\usepackage{algorithmic}
\usepackage{algorithm}
\usepackage{todonotes}
\usepackage{amsthm}
\usepackage{amsmath}
\usepackage{float}
\usepackage{caption}
\usepackage{xcolor}
\usepackage{subcaption}

\definecolor{darkblue}{rgb}{0.0, 0.0, 0.55}

\usepackage[colorlinks=true, allcolors=darkblue]{hyperref}

\newcommand{\dd}{{\mathrm{d}}}
\newcommand{\dw}{{\mathrm{w}}}
\newcommand{\dx}{{\mathbf{x}}}
\newcommand{\bbf}{{\mathbf{f}}}

\newtheorem{theorem}{Theorem}

\title{Recurrent Interpolants for Probabilistic Time Series Prediction}

\usepackage{amsmath,amsfonts,bm}

\newcommand{\diff}{{\mathrm{d}}}

\def\eqref#1{equation~\ref{#1}}
\def\Eqref#1{Equation~\ref{#1}}

\def\1{\bm{1}}

\def\rd{{\textnormal{d}}}

\def\vzero{{\bm{0}}}

\def\vmu{{\bm{\mu}}}

\def\mI{{\bm{I}}}

\DeclareMathAlphabet{\mathsfit}{\encodingdefault}{\sfdefault}{m}{sl}
\SetMathAlphabet{\mathsfit}{bold}{\encodingdefault}{\sfdefault}{bx}{n}

\newcommand{\E}{\mathbb{E}}

\newcommand{\R}{\mathbb{R}}

\author[1]{Yu Chen$^{*,}$}
\author[1]{Marin Bilo\v{s}$^{*,}$}
\author[2,3]{Sarthak Mittal$^{\dagger,*,}$}
\author[1]{Wei Deng}
\author[1]{Kashif Rasul}
\author[1]{Anderson Schneider}
\affil[1]{%
    Morgan Stanley 
}
\affil[2]{%
    Mila
}
\affil[3]{%
    Universit\'e de Montr\'eal
  }
\newcommand\blfootnote[1]{%
  \begingroup
  \renewcommand\thefootnote{}\footnote{#1}%
  \addtocounter{footnote}{-1}%
  \endgroup
}

\begin{document}

\maketitle

\begin{abstract}

Sequential models like recurrent neural networks and transformers have become standard for probabilistic multivariate time series forecasting across various domains. Despite their strengths, they struggle with capturing high-dimensional distributions and cross-feature dependencies. Recent work explores generative approaches using diffusion or flow-based models, extending to time series imputation and forecasting. However, scalability remains a challenge. This work proposes a novel method combining recurrent neural networks' efficiency with diffusion models' probabilistic modeling, based on stochastic interpolants and conditional generation with control features, offering insights for future developments in this dynamic field.

\end{abstract}
\blfootnote{Preprint}
\blfootnote{$^*$ Equal Contribution}
\blfootnote{$^\dagger$ Work done as part of an internship at Morgan Stanley}
\section{Introduction}

Autoregression models \citep{AR_models}, such as recurrent neural networks \citep{rnn, rnn2, lstm} or transformer models \citep{vaswani2017attention}, have been the go-to methods for neural time series forecasting. They are widely applied in finance, biological statistics, medicine, geophysical applications, etc., effectively showcasing their ability to capture short-term and long-term sequential dependencies \citep{NRDE}.
These methods can also provide an assessment of prediction uncertainty through probabilistic forecasting by incorporating specific parametric probabilistic models into the output layer of the neural network. For instance, a predictor can model the Gaussian distribution by predicting both mean and covariance. However, the probabilistic output layer is confined within a simple probability family because the density needs to be parameterized by neural networks, and the loss must be differentiable with respect to neural network parameters.

To better capture sophisticated distributions in time series modeling and learn both the temporal and cross-feature dependencies, a common strategy involves exploring the generative modeling of time series using efficient distribution transportation plans, especially via diffusion or flow-based models. 
For example, recent works such as \citet{scalable_SDE} propose using latent neural SDE as latent states for modeling time series in a stochastic manner, while
\citet{coupling_nkf} summarize non-linear extensions of state space models using both deterministic and stochastic transformation plans.
\citet{CSDI, marin_icml_23, chen2023provably, diffusion_state_space, generative_diffusion_denoise} studied the application of diffusion models in probabilistic time series imputation and forecasting. 
The generative model is trained to learn the joint density of the time series window $\mathbf{X}_{\mathrm{pred.}} \in \mathbb{R}^{D \times T_\mathrm{prediction}}$ with $D>1$ variates given $\mathbf{X}_{\mathrm{cont.}} \in \mathbb{R}^{D \times T_\mathrm{context}}$. $T_\mathrm{context}$ is the size of the context window and $T_\mathrm{prediction}$ is the size of the subsequent prediction window.
During inference, the model performs conditional generation given only the context, similar to the inpainting task in computer vision \citep{song2021sde}.
Compared to a recurrent model, where the model size is only proportional to the number of features, but not the length of the time window, such generative model predictors may suffer from scalability issues because the model size is related to both feature dimension and the size of the window.
A more computational-friendly framework is needed for large-scale generative model-based time series prediction problems.

Generative modeling excels at modeling complicated high-dimension distributions, but most models require learning a mapping from noise distribution to data distribution. 
If the generative procedure starts from an initial distribution proximate to the terminal data distribution, it can remarkably alleviate learning challenges, reduce inference complexity, and enhance the quality of generated samples, which is also supported by previous studies \citep{latent_ode, rasul2021autoregressive, rasul2020multivariate, chen2023schrodinger, rSB, VSDM, stochastic_interpolant_in_TS}.
Time series data is typically continuous and neighboring time points exhibit strong correlations, indicating that the distribution of future time points is close to that of the current time point.

These observations inspire the creation of a time series prediction model under the generative framework that maps between dependent data points:
initiating the prediction of future time point's distribution with the current time point is more straightforward and yields better quality;
meanwhile, the longer temporal dependency is encoded by a recurrent neural network and the embedded history is passed to the generative model as the guidance of the prediction for the future time points.
The new framework benefits from the efficient training and computation inherited from the recurrent neural network, while enjoying the high quality of probabilistic modeling empowered by the diffusion model.

Our contributions include:
\begin{itemize}
\item extending the theory of stochastic interpolants to a more general conditional generation framework with extra control features;
\item adopting a conditional stochastic interpolants module for the sequential modeling and multivariate probabilistic time series prediction tasks, which is computational-friendly and achieves high-quality modeling of the future time point's distribution.
\end{itemize}

\section{Background }  \label{sec:background}
As we formalize probabilistic time series forecasting within the generative framework in Section \ref{sec:methods}, this section is dedicated to reviewing commonly used generative methods and their extensions for conditional generation. These models will serve as baseline models in subsequent sections. For a broader overview of time series forecasting problems, refer to \cite{salinas2019high, alexandrov2019gluonts}, and the references therein.

\subsection{Denoising Diffusion Probabilistic Model (DDPM)}
DDPM \citep{sohl-dickstein15, ddpm} adds Gaussian noise to the observed data point $\mathbf{x}^0 \in \R^D$ at different scales, indexed by $n$, $0 < \beta_1 < \beta_2 < \dots < \beta_N$ such that the first noisy value $\mathbf{x}^1$ is close to the clean data $\mathbf{x}^0$, and the final value $\mathbf{x}^N$ is indistinguishable from noise. The generative model learns to revert this process allowing sampling new points from pure noise samples.

Following previous convention, we define $\bar{\alpha}_n = \prod_{k=1}^n \alpha_k$, with $\alpha_n = 1 - \beta_n$. Then when the transition kernel is Gaussian it can be computed directly from $\mathbf{x}^0$:
\begin{align}\label{eq:ddpm_q_xi_x0}
q(\mathbf{x}^n | \mathbf{x}^0) = \mathcal{N}(\sqrt{\bar{\alpha}_n} \mathbf{x}^0, (1 - \bar{\alpha}_n) \mI) .
\end{align}
The \emph{posterior} distribution is available in closed form:
\begin{align}\label{eq:ddpm_q_posterior}
q(\mathbf{x}^{n-1} | \mathbf{x}^n, \mathbf{x}^0) = \mathcal{N}(\tilde{\vmu}_n, \tilde\beta_n \mI),
\end{align}
where $\tilde{\vmu}_n$ depends on $\mathbf{x}^0$, $\mathbf{x}^n$ and a choice of $\beta$-scheduler.
The generative model $p(\mathbf{x}^{n-1} | \mathbf{x}^n) \approx q(\mathbf{x}^{n-1} | \mathbf{x}^n, \mathbf{x}^0)$ approximates the reverse process. The actual model $\epsilon_\theta(\mathbf{x}^0, n)$ is usually reparameterized to predict the noise added to a clean data point, from the noisy data point $\mathbf{x}^n$. The loss function can be simply written as:
\begin{align}\label{eq:ddpm_loss}
\mathcal{L} = 
    \E_{\bm{\epsilon}\sim\mathcal{N}(\vzero,\mI),n\sim\mathcal{U}(\{1,\dots,N\})}
    \left[\lVert 
    \epsilon_\theta(\mathbf{x}^n, n) - \bm{\epsilon}
    \rVert_2^2\right].
\end{align}
Sampling new data is performed by first sampling a point from the pure noise $\mathbf{x}^N \sim \mathcal{N}(\vzero, \mI)$ and then gradually denoising it using the above model to get a sample from the data distribution via $N$ calls of the model \citep{ddpm}.

\subsection{Score-based Generative Model (SGM)}
SGM \citep{song2021sde}, like DDPM, considers a pair of forward and backward dynamics between $s\in[0, 1]$:
\begin{align}
\diff \mathbf{x}^s  =& f(\mathbf{x}^s , s) \diff s + g(s) \diff \textbf{w}^s 
\label{eq:sgm_sde} \\
\diff \mathbf{x}^s  =& [f(\mathbf{x}^s , s) - g(s)^2 \nabla_{\mathbf{x}^s } \log p(\mathbf{x}^s )] \diff s + g(s) \diff \textbf{w}^s, \label{eq:sgm_reverse_sde}
\end{align}
where $\nabla_{\mathbf{x}^s } \log p(\mathbf{x}^s )$ is the so-called score function.
The forward process usually is scheduled as simple processes, such as Brownian motion or Ornstein–Uhlenbeck process, which can transport data distribution to standard Gaussian distribution.
The generative process is achieved by the backward process that walks from Gaussian prior distribution to the data distribution of interest.
Now, \Eqref{eq:sgm_reverse_sde} gives a way to generate new points by starting at $\mathbf{x}^1 \sim \mathcal{N}(\vzero, \mI)$ and solving the SDE backward in time giving $\mathbf{x}^0$ as a sample from data distribution. In practice, the only missing piece is obtaining the score. 
A standard approach is to approximate the score with a neural network.

Since during training, we have access to clean data, the score function is available in closed form. The model $\epsilon_\theta(\mathbf{x}^s, s)$ learns to approximate the score from noisy data only, resulting in a loss function similar to \Eqref{eq:ddpm_loss}:
\begin{align}\label{eq:sgm_loss}
\begin{split}
    \mathcal{L} =& 
        \E_{s\sim\mathcal{U}(0, 1), \mathbf{x}^0 \sim \text{Data},
            \mathbf{x}^s \sim p(\mathbf{x}^s | \mathbf{x}^0) }
        \Big[ \\
    &\lVert 
        \epsilon_\theta(\mathbf{x}^s , s) - \nabla_{\mathbf{x}^s }\log p(\mathbf{x}^s | \mathbf{x}^0)
    \rVert_2^2
    \Big] .
\end{split}
\end{align}

\subsection{Flow Matching (FM)}

Flow matching \citep{lipman2022flow} constructs a probability path by learning the vector field that generates it. Given a data point $\mathbf{x}^1$, the conditional probability path is denoted with $p_s(\mathbf{x} | \mathbf{x}^1)$ for $s \in [0, 1]$. We put the constraints on $p_s(\mathbf{x} | \mathbf{x}^1)$ such that $p_0(\mathbf{x} | \mathbf{x}^1) = \mathcal{N}(\vzero, \mI)$ and $p_1(\mathbf{x} | \mathbf{x}^1) = \mathcal{N}(\mathbf{x}^1, \sigma^2 \mI)$, with small $\sigma > 0$. That is, the distribution $p_0(\mathbf{x} | \mathbf{x}^1)$ corresponds to the noise distribution and the distribution $p_1(\mathbf{x} | \mathbf{x}^1)$ is centered around the data point with small variance.

Then there exists a conditional vector field $u_s(\mathbf{x} | \mathbf{x}^1)$ which generates $p_s(\mathbf{x} | \mathbf{x}^1)$. Our goal is to learn the vector field with a neural network $\epsilon_\theta(\mathbf{x}, s)$ which amounts to learning the generative process. This can be done by minimizing the flow matching objective:
\begin{align}\label{eq:fm_loss}
\begin{split}
    \mathcal{L} =& E_{s\sim\mathcal{U}(0, 1), \mathbf{x}^1 \sim \text{Data}, \mathbf{x} \sim p_s(\mathbf{x} | \mathbf{x}^0)}
    \Big[ \\
    &\lVert 
        \epsilon_\theta(\mathbf{x}, s) - u_s(\mathbf{x}^s | \mathbf{x}^1)
    \rVert_2^2
    \Big] .
\end{split}
\end{align}
Going back to \Eqref{eq:sgm_loss} we notice that the two approaches have similarities. Flow matching differs in the path constructions and it learns the vector field directly, instead of learning the score, potentially offering a more stable alternative.

One choice for the noising function is transporting the values into noise as a linear function of transport time:
\begin{align}
    \mathbf{x}^s = s \mathbf{x}^1 + (1 - (1 - \sigma) s) \mathbf{\epsilon}, \quad \mathbf{\epsilon} \sim \mathcal{N}(\vzero, \mI) .
\end{align}
The probability path is generated by the following conditional vector field which is available in closed form:
\begin{align}
    u_s(\mathbf{x} | \mathbf{x}^1) = \frac{\mathbf{x}^1 - (1 - \sigma) \mathbf{x}}{1 - (1 - \sigma) s} .
\end{align}
By learning the field $u_s(\mathbf{x} | \mathbf{x}^1)$ with a neural network $\epsilon_\theta(\mathbf{x}, s)$ we can sample new points by sampling an initial value $\mathbf{x}^0$ from the noise distribution $p_0$ and solve an ODE $0 \mapsto 1$ to obtain the new sample $\mathbf{x}^1$. 

\subsection{Stochastic Interpolants (SI)}

\begin{figure}[t]
\centering
\includegraphics[width=0.35\textwidth]{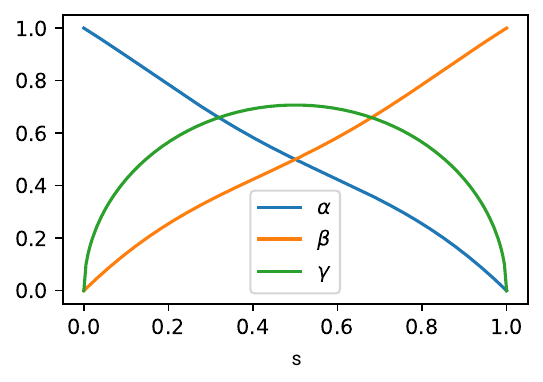}
\caption{$\alpha(\cdot)$, $\beta(\cdot)$,  and $\gamma(\cdot)$, the schedules of stochastic interpolants.}
\label{fig:si_schedule}
\end{figure}

Stochastic interpolants \citep{conditional_interpolant} aims to model the \textit{dependent couplings} between $(\mathbf{x}^0, \mathbf{x}^1)$  with their joint density $\rho(\mathbf{x}^0, \mathbf{x}^1)$, and establish a two-way generative SDEs mapping from one data distribution to another.
The method constructs a straightforward stochastic mapping from $s=0$ to $s=1$ given the values at two ends $\mathbf{x}^0 \sim \rho_0$ to $\mathbf{x}^1 \sim \rho_1$, which provides a means of transport between two densities $\rho_0$ and $\rho_1$, while maintaining the dependency between $\mathbf{x}^0$ and $\mathbf{x}^1$. 
\begin{equation}
\mathbf{x}^s = \alpha(s) \mathbf{x}^0 + \beta(s) \mathbf{x}^1 + \gamma(s) \mathbf{z}, 
\; s\in [0, 1], \mathbf{z}\sim \mathcal{N}(\mathbf{0}, \mathbf{I})
\label{eq:interpolant}
\end{equation}
where $\rho(s, \mathbf{x})$ is the marginal density of $\mathbf{x}^s$ at diffusion time $s$. 
Such a stochastic mapping is characterized by a pair of functions: velocity function $\mathbf{b}(s, \mathbf{x})$ and score function 
$\mathbf{s}(s, \mathbf{x})$:
\begin{gather}
\mathbf{s}(s, \mathbf{x}) := \nabla \log \rho(s, \mathbf{x}), \\
\mathbf{b}(s, \mathbf{x}) := \mathbb{E}_{\mathbf{x}^0, \mathbf{x}^1, \mathbf{z}}
[\dot{\alpha}(s) \mathbf{x}^0 
+ \dot{\beta}(s) \mathbf{x}^1 
+ \dot{\gamma}(s)\mathbf{z} 
    | \mathbf{x}^s= \mathbf{x} ].
\end{gather}
$\mathbf{b}(s, \mathbf{x})$, $\rho(s, \mathbf{x})$, and $\mathbf{s}(s, \mathbf{x})$ satisfy the equality below,
\begin{gather}
\partial_t \rho(s, \mathbf{x})
+ \nabla \cdot (\mathbf{b}(s, \mathbf{x}) \rho(s, \mathbf{x}))
=0 \\
\mathbf{s}(s, \mathbf{x})= -\gamma^{-1}(s)\mathbb{E}_{\mathbf{z}}
[\mathbf{z}  | \mathbf{x}^s = \mathbf{x}],
\end{gather}
where $\alpha(s)$ and $\beta(s)$ schedule the deterministic interpolant.
We set $\alpha(0)=1, \alpha(1)=0, \beta(0)=0, \beta(1)=1$.
$\gamma(s)$ schedules the variance of the stochastic component $\mathbf{z}$.
We set $\gamma(0) = \gamma(1) = 0$, so the two ends of the interpolant are fixed at $\mathbf{x}^0$ and $\mathbf{x}^1$.
Figure \ref{fig:si_schedule} shows one example of the interpolant schedule,
where $\alpha(s) = \sqrt{1 - \gamma^2(s)} \cos(\frac{1}{2} \pi s)$,
$\beta(s) = \sqrt{1 - \gamma^2(s)} \sin(\frac{1}{2} \pi s)$,
$\gamma(s) = \sqrt{2s(1-s)}$.

The velocity function $\mathbf{b}(s, \mathbf{x})$ and the score function $\mathbf{s}(s, \mathbf{x})$ can be modeled by a rich family of functions, such as deep neural networks. The model is trained to match the above equality by minimizing the mean squared error loss functions,
\begin{equation} \label{eq:si_b_loss}
\begin{aligned}
\mathcal{L}_b=&\int_0^1 \mathbb{E}\Big[
\frac{1}{2}\| \hat{\mathbf{b}} (s,\mathbf{x}^s) \|^2 \\
& - \big(\dot{\alpha}(s) \mathbf{x}^0 
+ \dot{\beta}(s) \mathbf{x}^1 
+ \dot{\gamma}(s)\mathbf{z} \big)^T \hat{\mathbf{b}}(s,\mathbf{x}^s)
\Big] \diff s
\end{aligned}
\end{equation}
\begin{equation} \label{eq:si_s_loss}
\begin{aligned}
\mathcal{L}_s=& \int_0^1 \mathbb{E}\Big[
\frac{1}{2}\| \hat{\mathbf{s}} (s,\mathbf{x}^s) \|^2 
 + \gamma^{-1}\mathbf{z}^T \hat{\mathbf{s}}(s,\mathbf{x}^s)
\Big] \diff s.
\end{aligned}
\end{equation}
More details of training will be shown in section \ref{sec:methods}.

During inference, usually, one side of the diffusion trajectory at $s=0$ or $s=1$ is given, the goal is to infer the sample distribution on the other side.
The interpolant in \eqref{eq:interpolant} results in elegant forward and backward SDEs and corresponding Fokker-Planck equations, which offer convenient tools for inference.
The SDEs are composed of $\mathbf{b}(s, \mathbf{x}^s)$ and $\mathbf{s}(s, \mathbf{x}^s)$, which are learned from the data.
For any $\epsilon(s) \geq 0$, define the forward and backward SDEs
\begin{align}
\diff\mathbf{x}^s =& [\mathbf{b}(s, \mathbf{x}) 
    + \epsilon(s)\mathbf{s}(s,\mathbf{x})] \diff s 
    + \sqrt{2\epsilon(s)} \diff \mathbf{w}^s  \label{eq:si_forward_sde} \\
\diff\mathbf{x}^s =& [\mathbf{b}(s, \mathbf{x}) 
    - \epsilon(s)\mathbf{s}(s,\mathbf{x})] \diff s 
+ \sqrt{2\epsilon(s)} \diff \mathbf{w}_{\mathrm{B}}^s,  \label{eq:si_backward_sde}
\end{align}
where $\mathbf{w}_{\mathrm{B}}^s$ is the backward Brownian motion.
The SDEs satisfy the forward and backward Fokker-Plank equations,
\begin{align}
& \partial_s \rho + \nabla\cdot (\mathbf{b}_{\mathrm{F}} \rho) = \epsilon(s)\Delta \rho, \rho(0) = \rho_0 \\
& \partial_s \rho + \nabla\cdot (\mathbf{b}_{\mathrm{B}} \rho) = -\epsilon(s)\Delta \rho, \rho(1) = \rho_1.
\end{align}
These properties imply that one can draw samples from the conditional density $\rho(\mathbf{x}^1 | \mathbf{x}^0)$ following the forward SDE in \eqref{eq:si_forward_sde} starting from $\mathbf{x_0}$ at $s=0$.
It can also draw samples from the joint density $\rho(\mathbf{x}^0, \mathbf{x}^1)$ by initially drawing a sample $\mathbf{x}^0 \sim \rho_0$ (if feasible, for example, pick one sample from the dataset), then using the forward SDE to generate samples $\mathbf{x}^1$ at $s=1$. 
The method guarantees that $\mathbf{x}^1$ follows marginal distribution $\rho_1$ and the sample pair $(\mathbf{x}^0, \mathbf{x}^1)$ satisfies the joint density $\rho(\mathbf{x}^0, \mathbf{x}^1)$.
Drawing samples using the backward SDE is similar: one can draw samples from $\rho(\mathbf{x}^0 | \mathbf{x}^1)$ and the joint density $\rho(\mathbf{x}^0, \mathbf{x}^1)$ as well.
Details of inference will be shown in section \ref{sec:methods}.

\begin{figure}[t]
\centering
\includegraphics[width=60mm]{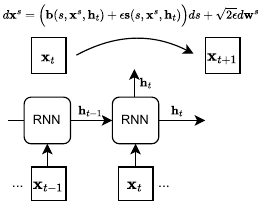}
\caption{Stochastic interpolants for time series prediction using forward SDE in \eqref{eq:si_forward_sde_cond}. 
}
\label{fig:si_diagram}
\end{figure}
\section{Conditional generation with extra features}
All the aforementioned methods can be adapted for conditional generation with additional features. The conditions may range from simple categorical values \citep{song2021sde} to complex prompts involving multiple data types, including partial observations of a sample's entries (e.g., image inpainting, time series imputation) \citep{CSDI, song2021sde}, images \citep{zheng2023layoutdiffusion, rombach2022high}, text \citep{rombach2022high, zhang2303text}, etc. A commonly employed technique to handle diverse conditions is to integrate condition information through feature embedding, where the embedding is injected into various layers of neural networks \citep{song2021sde, rombach2022high}. For instance, conditional SGM can be trained with
\begin{equation} \label{eq:sgm_cond_loss}
\begin{aligned}
\mathcal{L}_{\mathrm{cond}} =& 
    \E_{s\sim\mathcal{U}(0, 1), (\mathbf{x}^0, \xi) \sim \text{Data},  
        \mathbf{x}^s \sim p(\mathbf{x}^s|\mathbf{x}^0)}
    \Big[ \\
&\lVert 
    \epsilon_\theta(s, \mathbf{x}^s, \xi) - \nabla_{\mathbf{x}^s}\log p(\mathbf{x}^s| \mathbf{x}^0)
\rVert_2^2
\Big].
\end{aligned}
\end{equation}
where the data is given by pairs of a sample $\mathbf{x}^0$ and the corresponding condition $\xi$.
This simple scheme approach showcases its effectiveness in various tasks, achieving state-of-the-art performance \citep{rombach2022high, zhang2303text}.

\begin{figure}[h]
\centering
\includegraphics[width=0.45\textwidth]{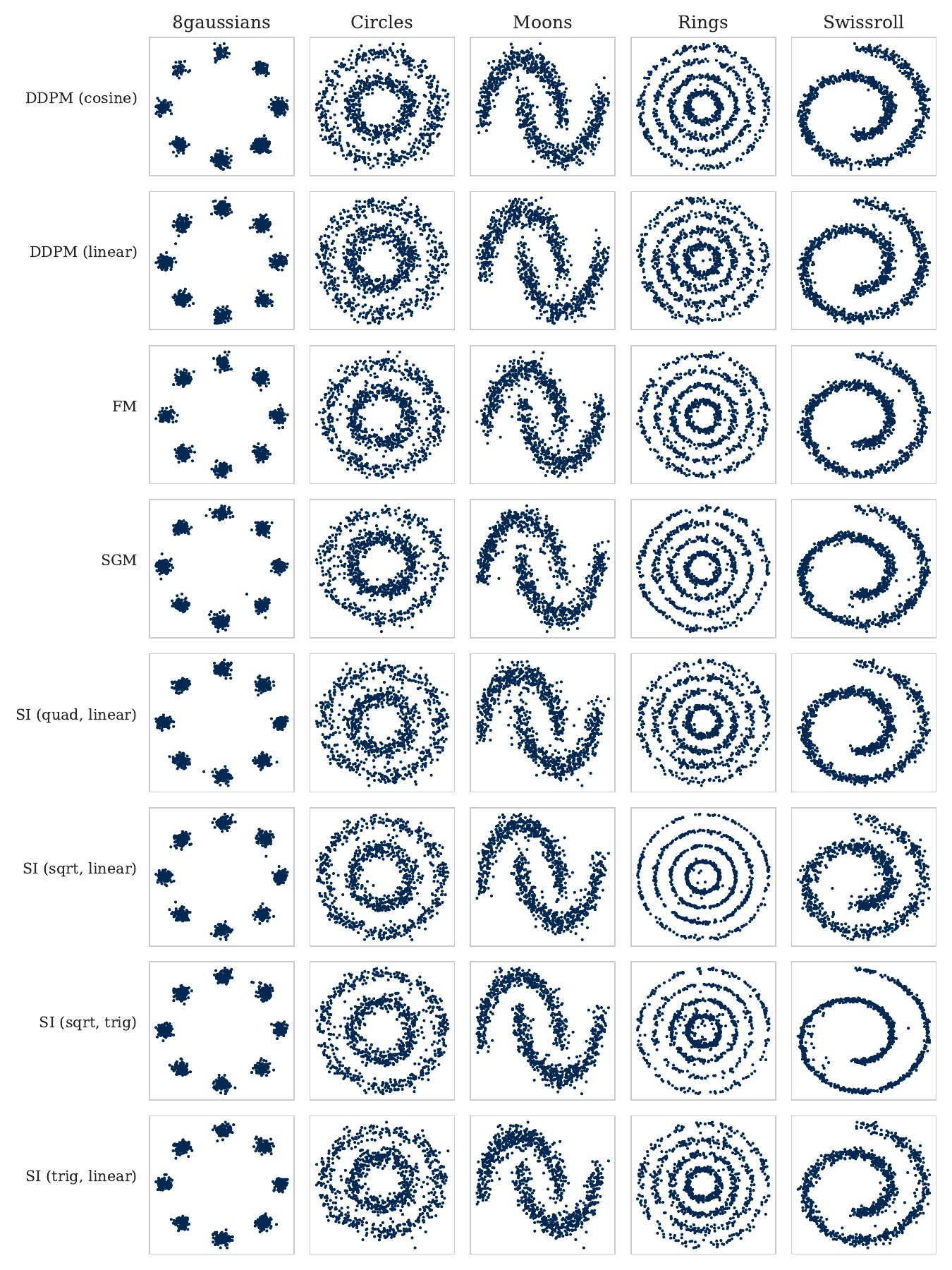}
\caption{Examples of model generated samples for synthetic two-dimensional ($D=2$) datasets.}
\label{fig:synthetic_experiment}
\end{figure}
Likewise, SI can be expanded for conditional generation by substituting the velocity function and score function with $\mathbf{b}(\mathbf{x}^s, s, \xi)$ and $\mathbf{s}(\mathbf{x}^s, s, \xi)$ \citep{conditional_interpolant}. The model is trained using samples of tuples $(\mathbf{x}^0, \mathbf{x}^1, \xi)$, where $\xi$ is the extra condition feature.
Consequently, the inference using forward or backward SDEs becomes
\begin{align}
\diff\mathbf{x}^s =& [\mathbf{b}(s, \mathbf{x}^s, \xi)
+ \epsilon(s) \mathbf{s}(s, \mathbf{x}^s, \xi)]\diff s + \sqrt{2\epsilon(s)} \diff\mathbf{w}^s  
\label{eq:si_forward_sde_cond} \\
\diff\mathbf{x}^s =& [\mathbf{b}(s, \mathbf{x}^s, \xi)
- \epsilon(s) \mathbf{s}(s, \mathbf{x}^s, \xi)]\diff s + \sqrt{2\epsilon(s)} \diff\mathbf{w}_{\mathrm{B}}^s,
\label{eq:si_backward_sde_cond}
\end{align}
where both velocity and score functions depend on the condition $\xi$.
The loss functions are similar to \eqref{eq:si_b_loss} and \eqref{eq:si_s_loss}.

Regarding the time series prediction task, we will encode a large context window as the conditional information, and the prediction or generation of future time points will rely on such a conditional generation mechanism.

Next, we demonstrate that the probability distribution of $\mathbf{x}^s$ as simulated by \eqref{original_main}, results in a dynamic density function. This density serves as a solution to a transport equation \ref{dyn_transport}, which smoothly transitions between $\rho_0$ and $\rho_1$. 
\begin{algorithm}[t]
   \caption{Training algorithm.}
   \label{alg:si_training}
\begin{algorithmic}
\STATE {\bfseries Input:} Sample $\mathbf{x}_{1:C+P}$ from training split. Interpolant schedules: $\alpha(s), \beta(s), \gamma(s)$.
Models: velocity $\hat{\mathbf{b}}$, score $\hat{\mathbf{s}}$, $\mathrm{RNN}$. 
\FOR{iteration $t = C$ {\bfseries to} $C+P-1$}
\STATE
$s \sim \mathrm{Beta}(0.1, 0.1)$ and $\mathbf{z} \sim \mathcal{N}(\mathbf{0}, \mathbf{I})$. \\
$\mathbf{x}^{s} = \alpha(s) \mathbf{x}_{t} + \beta(s) \mathbf{x}_{t+1} + \gamma(s) \mathbf{z}$ \\
$\mathbf{h}_t = \mathrm{RNN}(\mathbf{x}_{t}, \mathbf{h}_{t-1})$
\begin{equation*}
\begin{aligned}
&\mathcal{L}_b=  \frac{1}{p_{\mathrm{Beta}}(s)} \Big[
\frac{1}{2}\| \hat{\mathbf{b}} (s, \mathbf{x}^{s}, \mathbf{h}_t) \|^2 \\
& - \big(\dot{\alpha}(s) \mathbf{x}_{t}
+ \dot{\beta}(s) \mathbf{x}_{t+1}
+ \dot{\gamma}(s)\mathbf{z} \big)^T \hat{\mathbf{b}}(s,\mathbf{x}^{s}, \mathbf{h}_t) \Big] \\
& \mathcal{L}_s =  \frac{1}{p_{\mathrm{Beta}}(s)} \Big[
\frac{1}{2}\| \hat{\mathbf{s}} (s, \mathbf{x}^{s}) \|^2 
+ \gamma^{-1}\mathbf{z}^T \hat{\mathbf{s}}(s, \mathbf{x}^{s})
\Big]
\end{aligned}
\end{equation*} \\
Perform back-propagation by minimizing $\mathcal{L}_b$ and $\mathcal{L}_s$.
\ENDFOR
\end{algorithmic}
\end{algorithm}

\begin{theorem}\label{SI_theorem}
(\textit{Extension of Stochastic Interpolants to Arbitrary Joint Distributions}). Let $\rho_{01}$ be the joint distribution $(\mathbf{x}^0, \mathbf{x}^1) \sim \rho_{01}$ and let the stochastic interpolant be 
\begin{align}
\mathbf{x}^s = \alpha_s \mathbf{x}^0 + \beta_s \mathbf{x}^1 + \gamma_s \mathbf{z},\label{original_main}
\end{align}
where $\alpha_0=\beta_1=1$, $\alpha_1=\beta_0=\gamma_0 = \gamma_1 = 0$, and $\alpha_s^2+\beta_s^2+\gamma_s^2>0$ for all $s\in[0,1]$. We define $\rho_s$ to be the noise-dependent density of $\mathbf{x}^s$, which satisfies the boundary conditions at $s = 0, 1$ and the transport equation follows that
\begin{align}
\dot{\rho}_s + \nabla \cdot (\mathbf{b}_s \rho_s) = 0\label{dyn_transport}
\end{align}
for all $s \in[0, 1]$ with the velocity defined as
\begin{align}
\mathbf{b}_s(\mathbf{x}|\xi) = \mathbb{E}\left[\dot{\alpha_s} \mathbf{x}^0 + \dot{\beta_s} \mathbf{x}^1 + \dot{\gamma}_s\, \mathbf{z} |\mathbf{x}^s=\mathbf{x}, \xi \right],
\end{align}
where the expectation is based on the density $\rho_{01}$ given $\mathbf{x}^s=\mathbf{x}$ and the extra information $\xi$.

The score function follows the relation such that 
$$\nabla \log \rho_s(\mathbf{x}) = -\gamma^{-1}_s \mathbb{E}\left[\mathbf{z} | \mathbf{x}^s = \mathbf{x}, \xi\right].$$
\end{theorem}

The proof is in a spirit similar to Theorem 2 in \cite{conditional_interpolant} and detailed in section \ref{cond_SI}. The key difference is that we consider a continuous-time interpretation and avoid using characteristic functions, which makes the analysis more friendly to users. Additionally, the score function $\nabla \log \rho_s(\mathbf{x})$ is optimized in a simple quadratic objective function as indicated in Theorem \ref{minimizers} in the Appendix. 

\section{Stochastic interpolants for time series prediction } \label{sec:methods}

We formulate the multivariate probabilistic time series prediction tasks through the conditional probability $\Pi_{t=C+1}^{C+P} p(\mathbf{x}_{t} | \mathbf{x}_{1:t-1})$ for some chosen context length $C$ and prediction length $P$. The model diagram is illustrated in Figure \ref{fig:si_diagram}. Here, $\mathbf{x}_{t} \in \mathbb{R}^D$ represents the multivariate time series at \emph{date-time index} $t$ with $D > 1$ variates.   $\mathbf{x}_{1:C} = \mathbf{X}_{\mathrm{cont.}}$ is the context window and during training the subsequent prediction window  $\mathbf{x}_{C+1:C+P} = \mathbf{X}_{\mathrm{pred.}}$ is available, typically sampled randomly from within the train split of a dataset.

For this problem, we employ the conditional Stochastic Interpolants (SI) method as follows.
In the \textit{training phase}, the generative model learns the joint distribution $p(\mathbf{x}_{t+1}, \mathbf{x}_{t} |  \mathbf{x}_{1:t-1})$ of the pair $(\mathbf{x}_{t+1}, \mathbf{x}_{t})$ given the past observations $\mathbf{x}_{1:t-1}$,
where $\mathbf{x}_{t} \sim \rho_0$ and  $\mathbf{x}_{t+1} \sim \rho_1$ for all $t$, so the marginal distributions are equal: $\rho_0 = \rho_1$.
The model aims to learn the coupling relation between $\mathbf{x}_{t+1}$ and $\mathbf{x}_{t}$ conditioning on the history $\mathbf{x}_{1:t}$.
This is achieved by training the conditional velocity and score functions in \eqref{eq:si_forward_sde_cond}.

\begin{algorithm}[t]
   \caption{Inference algorithm.}
   \label{alg:si_inference}
\begin{algorithmic}
\STATE {\bfseries Input:}
Last context $\mathbf{x}_{1:C}$. 
Trained models: Velocity $\hat{\mathbf{b}}$, score $\hat{\mathbf{s}}$, $\mathrm{RNN}$.
Diffusion variance $\epsilon(s)$. \\
\FOR{iteration $t = C$ {\bfseries to} $C+P-1$}
\STATE
Set $\hat{\mathbf{x}}^0 = \mathbf{x}_{t}$ and $\mathbf{h}_t = \mathrm{RNN}(\mathbf{x}_{t}, \mathbf{h}_{t-1} )$. \\
Run SDE integral for $s \in [0, 1]$ following
\begin{equation*}
\diff \hat{\mathbf{x}}^s = [\hat{\mathbf{b}}(s, \hat{\mathbf{x}}^s, \mathbf{h}_t)
+ \epsilon(s) \hat{\mathbf{s}}(s, \hat{\mathbf{x}}^s, \mathbf{h}_t)]\diff s 
+ \sqrt{2\epsilon} \diff \mathbf{w}^s
\end{equation*}
\STATE {\bfseries Output:} $\hat{\mathbf{x}}^1$ as prediction: $\mathbf{x}_{t+1}$.
\ENDFOR
\end{algorithmic}
\end{algorithm}

\begin{table}[bh]
\centering
\def\arraystretch{1.25}
\resizebox{\linewidth}{!}{%
    \begin{tabular}{lccccc}
    \toprule
         & 8gaussians & Circles & Moons & Rings & Swissroll \\
        \midrule
        DDPM (cosine) & 2.58 & 0.20 & 0.20 & 0.12 & 0.24 \\
        DDPM (linear) & 0.70 & 0.18 & 0.12 & \bf{0.11} & \bf{0.14} \\
        SGM & 1.10 & 0.30 & 0.35 & 0.32 & \bf{0.14} \\
        FM & 0.58 & 0.10 & \bf{0.11} & 0.09 & 0.15 \\
        SI (quad, linear) & \bf{0.52} & 0.15 & 0.32 & 0.12 & 0.16 \\
        SI (sqrt, linear) & 0.59 & 0.29 & 0.51 & 0.22 & 0.37 \\
        SI (sqrt, trig) & 0.75 & 0.25 & 0.50 & 0.48 & 0.36 \\
        SI (trig, linear) & \bf{0.52} & \bf{0.13} & 0.29 & 0.21 & 0.16 \\
        \bottomrule
    \end{tabular}
}
\caption{Wasserstein distance between the generated samples and true data.}
\label{tab:synthetic_metrics}
\end{table}

As the sample spaces of $\rho_0$ and $\rho_1$ must be the same, the generative model can not directly map the whole context window $\mathbf{x}_{1:t}$ to the target $\mathbf{x}_{t+1}$ due to different tensor sizes. Instead, a recurrent neural network is used to encode the context $\mathbf{x}_{1:t}$ into a \textit{history prompt} $\mathbf{h}_t \in \mathbb{R}^H$ vector.
Subsequently, the score function and velocity function perform conditional generation 
diffusing from $\mathbf{x}_{t}$ with the condition input $\mathbf{h}_t$ following \eqref{eq:si_forward_sde_cond}.

The training loss consists of tuples $(\mathbf{x}_{t+1}, \mathbf{x}_{t}, \mathbf{h}_t)$ for each time step $t$.
It is worth noting that the loss values become larger when $s$ is close to the two ends.
To address this, importance sampling is leveraged to better handle the integral over diffusion time in the loss functions \eqref{eq:si_b_loss} and \eqref{eq:si_s_loss} to stabilize the training, where we use Beta distribution for our proposal distribution. The algorithm is outlined in Algorithm \ref{alg:si_training}. Additional details can be found in Appendix \ref{appendix:experiment}.

In the \textit{inference phase}, the RNN first encodes the context $\mathbf{x}_{1:t}$ into the history prompt $\mathbf{h}_t$, then SI transports the context vector $\mathbf{x}_{t}$ to the target distribution with the condition $\mathbf{h}_t$,  following the forward SDE. Regarding the multiple-step prediction, we recursively run the step-by-step prediction in an autoregressive manner as outlined in Algorithm \ref{alg:si_inference}. By repeating the inference loop (in the batch dimension) we can obtain empirical samples from the predicted distribution which are used to quantify uncertainty.

\begin{figure}[t]
\centering
\includegraphics[width=\linewidth]{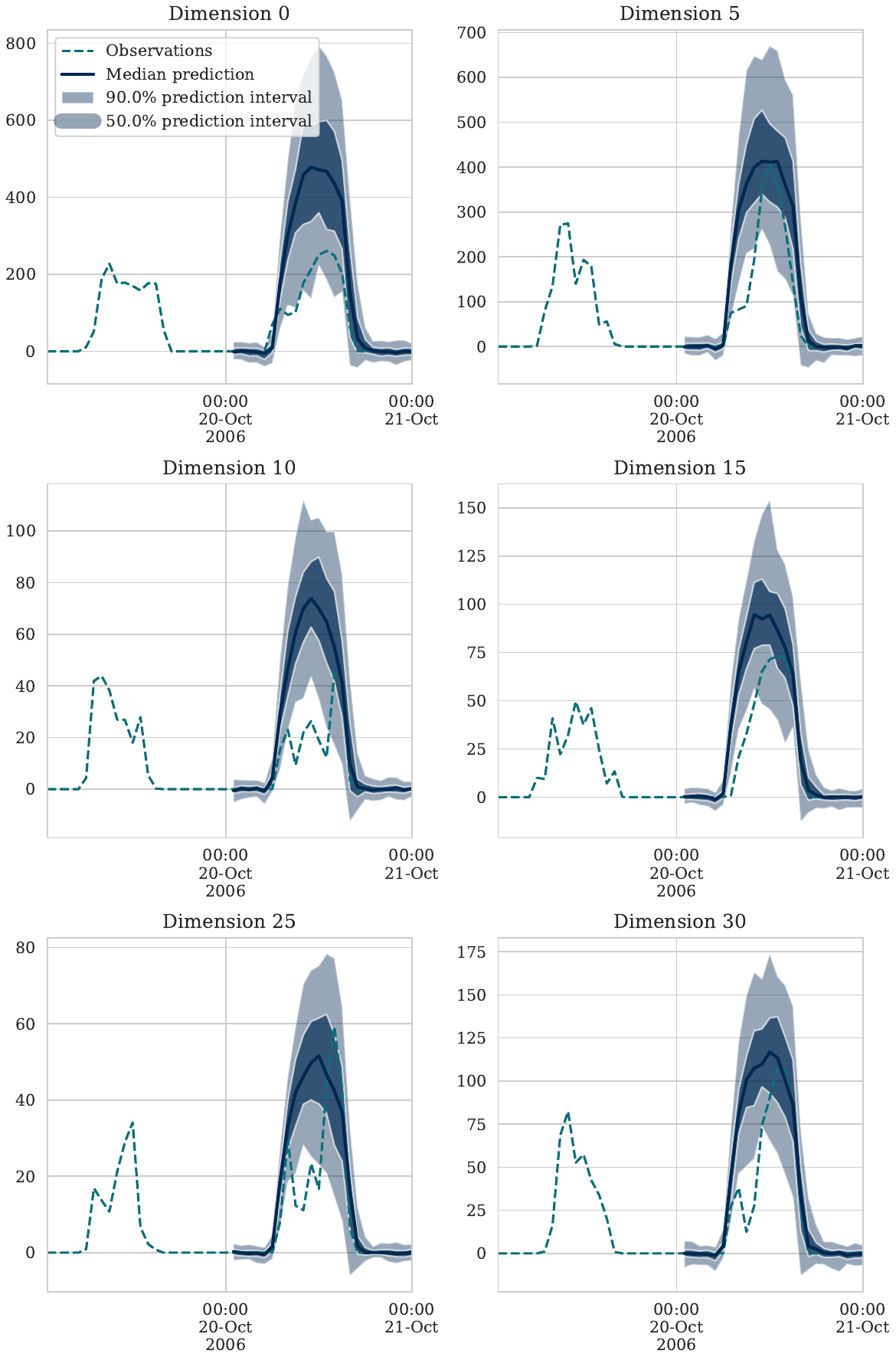}
\caption{Forecast paths for SI on Solar dataset showing median prediction, 50\textsuperscript{th} and 90\textsuperscript{th} confidence intervals calculated from model samples, on $6 \;/\; 137$ variate dimensions.}
\vspace{-2mm}
\label{fig:solar_forecasts}
\end{figure}

\section{Experiments}
We first verify the method on synthetic datasets and then apply it to the time series forecasting tasks with real data.

\begin{figure}[t]
\centering
\includegraphics{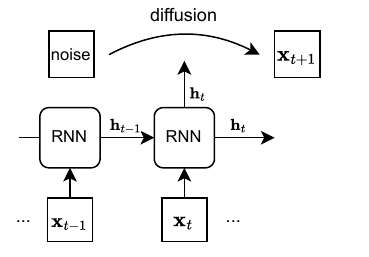}
\caption{Time-Grad \citep{rasul2021autoregressive} model for conditional time series prediction as a comparison.}
\label{fig:baseline_diagram}
\end{figure}
Baseline models such as DDPM, SGM, FM, and SI all involve modeling field functions, where the inputs are the state vector (in the same space of the data samples), diffusion time, and condition embedding, and the output is the generated sample.
The field functions correspond to the noise prediction function in DDPM;
the score function in SGM;
the vector field in FM;
and the velocity and score functions in SI. To make a fair comparison between these models, we use the same neural networks for these models.
Details of the models are discussed in Appendix \ref{appendix:experiment}.

\begin{table*}[t]
\centering
\def\arraystretch{1.25}
\begin{tabular}{lcccc}
\toprule
 & Exchange rate & Solar & Traffic & Wiki \\
\midrule
Vec-LSTM & 0.008\scriptsize{$\pm$0.001} & 0.391\scriptsize{$\pm$0.017} & 0.087\scriptsize{$\pm$0.041} & 0.133\scriptsize{$\pm$0.002} \\
DDPM & 0.009\scriptsize{$\pm$0.004} & \textbf{0.359}\scriptsize{$\pm$0.061} & 0.058\scriptsize{$\pm$0.014} & 0.084\scriptsize{$\pm$0.023} \\
FM & 0.009\scriptsize{$\pm$0.001} & 0.419\scriptsize{$\pm$0.027} & \textbf{0.038}\scriptsize{$\pm$0.002} & 64.256\scriptsize{$\pm$62.596} \\
SGM & 0.008\scriptsize{$\pm$0.002} & 0.364\scriptsize{$\pm$0.029} & 0.071\scriptsize{$\pm$0.05} & 0.108\scriptsize{$\pm$0.026} \\
SI & \textbf{0.007}\scriptsize{$\pm$0.001} & \textbf{0.359}\scriptsize{$\pm$0.06} & 0.083\scriptsize{$\pm$0.005} & \textbf{0.080}\scriptsize{$\pm$0.007} \\
\bottomrule
\end{tabular}
\caption{CRPS-sum metric on multivariate probabilistic forecasting tasks. A smaller number indicates better performance.}
\label{tab:crps_sum}
\end{table*}

\subsection{Synthetic datasets }

We synthesize several two-dimensional datasets with regular patterns, such as Circles, Moons, etc.
Details can be found in \cite{chen2021likelihood, lipman2022flow} and their published code repositories.
Models introduced in section \ref{sec:background} are compared to the SI as baselines.
For diffusion-like models, we implement DPPM with a linear or cosine noise scheduler.
We explore on synthetic datasets to determine a \emph{good} range of hyperparameters which will be used in later time series experiments. This experiment is used to investigate the properties with respect to the varying data sizes, model sizes, and training lengths.

To fairly compare the generation quality, all models are assigned to generate data in the same setting by mapping from standard Gaussian to the target distribution.
The neural networks and hyperparameters are also set as the same, such as batch size, training epochs, etc.
The generated samples from different methods are shown in Figure~\ref{fig:synthetic_experiment}.
Table~\ref{tab:synthetic_metrics} measures the sample quality with Wasserstein distance \citep{ramdas2017wasserstein}. It shows that all the models can capture the true distribution. 
The same holds when we use different metrics such as Sliced Wasserstein Distance (SWD) \citep{rabin2012wasserstein} and Maximum Mean Discrepancy (MMD) \citep{gretton2012kernel}.

We also test out different schedulers for a stochastic interpolant model. For example, ``SI (sqrt, linear)'' means we use square root gamma-function $\gamma(s) = \sqrt{2 s (1 - s)}$  and a linear interpolant. Other gamma-functions that we consider are \emph{quad}: $\gamma(s) = s (1-s)$, and \emph{trig}: $\gamma(s) = \sin^2(\pi s)$. We show that most of the gamma-interpolant function combinations achieve good results in modeling the target distribution.

\subsection{Multivariate probabilistic forecasting}

In this section, we will empirically verify that: 
1) SI is a suitable generative module for the prediction compared with other baselines with different generative methods under the same framework;
2) the whole framework can achieve competitive performance in time series forecasting.

\paragraph{Models.}
The baseline models include DDPM, SGM, and FM-based generative models adopted for step-by-step (autoregressive) prediction.
DDPM and SGM-based models can only generate samples by transporting Gaussian noise distribution to data distribution. So we modify the framework by replacing the context time point $\mathbf{x}_{t}$ with Gaussian noise, as shown in Figure \ref{fig:baseline_diagram}.
Flow matching can easily fit into this framework by replacing the denoising objective with the flow matching objective.
The modified framework is shown in Figure \ref{fig:si_diagram}. We model the map from the previous time series observation to the next (forecasted) value. We argue this is a more natural choice than mapping from noise for each time series prediction step.
Finally, Vec-LSTM from \cite{salinas2019high} is compared as a pure recurrent neural network model whose probabilistic layer is a multivariate Gaussian.

\paragraph{Setup.}
The real-world time series datasets include Solar \citep{lai2018modeling}, Exchange \citep{lai2018modeling}, 
Traffic\footnote{\url{https://archive.ics.uci.edu/ml/datasets/PEMS-SF}},  
and Wiki\footnote{\url{https://github.com/mbohlkeschneider/gluon-ts/tree/mv_release/datasets}}
which have been commonly used for probabilistic forecasting tasks.
We follow the preprocessing steps as in \cite{salinas2019high}.
The probabilistic forecasting is evaluated by Continuous Ranked Probability Score (CRPS-sum) \citep{koochali2022random}, normalized root mean square error via the median of the samples (NRMSE), 
and point-metrics normalized deviance (ND). 
The metrics calculation is provided by \texttt{gluonts} package \citep{alexandrov2019gluonts}.
In all of the cases, smaller values indicate better performance. 

\paragraph{Results.}

The results for CRPS-sum are shown in Table~\ref{tab:crps_sum}. The results for other metrics are consistent with CRPS-sum and are shown in Tables~\ref{tab:nd_sum} and \ref{tab:nrmse_sum}, in Appendix~\ref{appendix:experiment}. We outperform or match other models on three out of four datasets, only on Traffic FM model achieves better performance. Note that on Wiki data FM cannot capture the data distribution. We ran a search over flow matching hyperparameters without being able to get satisfying results. Therefore, we conclude that stochastic interpolants are a strong candidate for conditional generation, in particular for multivariate probabilistic forecasting.
By comparing to the RNN-based model Vec-LSTM, our model and other baselines such as SGM and DDPM get better performance, which implies that carefully modeling the probability distribution is critical for large dimension time series prediction.
Figure~\ref{fig:solar_forecasts} demonstrates the quality of the forecast on the Solar dataset. We can see that our model can make precise predictions and capture the uncertainty, even when the scale of the different dimensions varies considerably.

\section{Conclusions}

This study presents an innovative method that effectively merges the computational efficiency of recurrent neural networks with the high-quality probabilistic modeling of the diffusion model, specifically applied to probabilistic time series forecasting. Grounded in stochastic interpolants and an expanded conditional generation framework featuring control features, the method undergoes empirical evaluation on both synthetic and real datasets, showcasing its compelling performance.

\clearpage

\bibliography{uai.bib}

\clearpage
\appendix
\onecolumn

\section{Related Works}

A plethora of papers focus on auto-regression models, particularly transformer-based models. For a more comprehensive review, we refer to \cite{trans_survey, long_seq_ts}. While our work does not aim to replace RNN- or transformer-based architectures, we emphasize that one of the main motivations behind our work is to develop a probabilistic module building upon these recent advancements. Due to limited resources, we did not extensively explore all underlying temporal architectures but instead selected relatively simpler models as defaults.

The authors were aware of other diffusion-based probabilistic models, as highlighted in the introduction. Unlike our lightweight model, which models the transition between adjacent time points, these selected works model the entire time window, requiring both high memory and computational complexity. With our computation budget restricted to a 32 GB GPU device, effectively training these diffusion models on large datasets with hundreds of features is challenging.

Additionally, several relevant works are related to our idea. For instance, \cite{rasul2020multivariate} incorporates the DDPM structure, aligning with our DDPM baseline structure. During inference, the prediction diffuses from pure noise to the target distribution. TimeDiff \citep{time_diffusion} introduces two modifications to the established diffusion model: during training it mixes target and context data and it adds an AR model for more precise initial prediction. Both of these can be incorporated into our model as well.

The existing probabilistic forecasters model the distribution of the next value from scratch, meaning they start with normal distribution in the case of normalizing flows and diffusion models or output parametric distribution in the case of transformers and deep AR. We propose modeling the transformation between the previously observed value and the next value we want to predict. We believe this is a more natural way to forecast which can be seen from requiring fewer solver steps to reach the target distribution.

The second row (DDPM) in Table \ref{tab:crps_sum} is an exact implementation of \cite{rasul2020multivariate}. The results might be different due to slightly different training setups but all the models share the same training parameters so the rank should remain the same. We also include ND-sum and NRMSE-sum in the appendix for completeness.

\paragraph{Discussion of Vec-LSTM baseline} In terms of the neural network architecture, we use a similar architecture for the LSTM encoder. But to be clear, Vec-LSTM \citep{salinas2019high} and our SI framework are not the same mainly due to different ways of probabilistic modeling. Vec-LSTM considers the multivariate Gaussian distribution for the time points, where the mean and covariance matrices are modeled using separate LSTMs. Especially, the covariance matrix is modeled through a low-dimensional structure $\bm{\Sigma}(\bm{h}_t)=\bm{D}_t(\bm{h}_t)+\bm{V}_t(\bm{h}_t)\bm{V}_t(\bm{h}_t)^\intercal$, where $\bm{h}_t$
 is the latent variable from LSTM. The SI framework does not explicitly model the output distribution in any parametric format. Instead, the latent variable from RNN output is used as the condition variable to guide the diffusion model in Eq.\ref{eq:si_forward_sde_cond}. Thus, the architectures of RNNs in the two frameworks are not quite strictly comparable.

\section{Proof: Conditional Stochastic Interpolant}
\label{cond_SI}

The proof is in spirit similar to Theorem 2 in \cite{conditional_interpolant}. The key difference is that we consider a continuous-time interpretation, which makes the analysis more friendly to users.

\begin{proof}[Proof of Theorem \ref{SI_theorem}]

Given the conditional information $\xi$ and $\dx^s=\dx$ simulated from \eqref{original_main}, the conditional stochastic interpolant for \eqref{original_main} follows that (where the index $t$ is over the noise index and \emph{not} date-times):
\begin{align}
    \E[\dx^t|\dx^s=\dx, \xi]=\E[\alpha_t \dx^0+\beta_t \dx^1 + \gamma_t \mathbf{z} |\dx^s=\dx, \xi],\label{discrete_time}
\end{align}
where the expectation takes over the density for $(\dx^0, \dx^1)\sim \rho(\dx_0, \dx_1|\xi)$, $\xi\sim \eta(\xi)$, and $ \mathbf{z} \sim \mathcal{N}(\mathbf{0}, \mathbf{I})$.

We next show \eqref{discrete_time} is a solution of a stochastic differential equation as follows
\begin{align}
    \rd \E[\dx^t|\dx^s=\dx, \xi]&=\bbf_t(\dx) \dd t + \sigma_t\rd \dw^t,\label{continous_sde}
\end{align}

where $\bbf_t(\dx)=\E[\dot{\alpha_t} \dx^0+\dot{\beta_t} \dx^1|\dx^s=\dx, \xi]$ and $\sigma_t=\sqrt{2{\gamma_t} \dot{\gamma_t}}$. %

To prove the above argument, we proceed to verify the \emph{drift} and \emph{diffusion} terms respectively:
\begin{itemize}
    \item \emph{Drift}: It is straightforward to verify the drift $\bbf_t$ by taking the gradient of the conditional expectation $\E[\alpha_t \dx^0+\beta_t \dx^1|\dx^s=\dx, \xi]$ with respect to $t$. 

    \item \emph{Diffusion}: For the diffusion term, the proof hinges on showing $\sigma_t=\sqrt{2{\gamma_t} \dot{\gamma_t}}$, which boils down to prove the stochastic calculus follows that $\int_0^t \sqrt{2{\gamma_s} \dot{\gamma_s}} \rd \dw^s=\gamma_t \mathbf{z}$. Note that $\mathbb{E}[\int_0^t \sqrt{2{\gamma_s} \dot{\gamma_s}} \rd \dw^s]=0$. Invoking the It\^o isometry, we have $\text{Var}(\int_0^t \sqrt{2{\gamma_s} \dot{\gamma_s}} \rd \dw^s)=\int_0^t  2{\gamma_s} \dot{\gamma_s} \dd s=\int_0^t  ({\gamma_s}^2)' \dd s=\gamma_t^2$ (given $\gamma_0=0$). In other words, $\int_0^t \sqrt{2{\gamma_s} \dot{\gamma_s}} \rd \dw^s$ is a normal random variable with mean 0 and variance $\gamma_t^2$, which proves that \eqref{discrete_time} is a solution of the stochastic differential equation \ref{continous_sde}.
\end{itemize}

Define $\Sigma_t=2{\gamma_t} \dot{\gamma_t}$, we know the Fokker-Planck equation associated with \eqref{continous_sde} follows that

\begin{equation}
\begin{split}\label{FPK_Eqn}
    0&=\frac{\partial \rho_t}{\partial t}+\nabla \cdot \bigg(\bbf_t \rho_t - \frac{1}{2}\Sigma_t\nabla \rho_t\bigg) \\
    &=\frac{\partial \rho_t}{\partial t}+\nabla \cdot \bigg(\bigg(\bbf_t - \frac{1}{2}\Sigma_t\nabla \log \rho_t \bigg)\rho_t\bigg) \\
    &=\frac{\partial \rho_t}{\partial t}+\nabla \cdot \bigg(\bigg(\E[\dot{\alpha_t} \dx^0+\dot{\beta_t} \dx^1|\dx^s=\dx, \xi] - {\gamma_t} \dot{\gamma_t}\nabla \log \rho_t \bigg)\rho_t\bigg)\\
    &=\frac{\partial \rho_t}{\partial t}+\nabla \cdot \big(b_{t|s}(\dx, \xi)  \rho_t\big),
\end{split}
\end{equation}
where $b_{t|s}(\dx|\xi)=\E[\dot{\alpha_t} \dx^0+\dot{\beta_t} \dx^1- {\gamma_t} \dot{\gamma_t}\nabla \log \rho_t|\dx^s=\dx, \xi] $.

Further setting $s=t$ and rewrite $b_{t}\equiv b_{t|t}$, we have $b_{t}(\dx|\xi)=\E[\dot{\alpha_t} \dx^0+\dot{\beta_t} \dx^1- {\gamma_t} \dot{\gamma_t}\nabla \log \rho_t|\dx^t=\dx, \xi]$.

Further define $g^{(i)}_t(\dx|\xi)=\E[\dx^i|\dx^t=\dx, \xi]$, where $i\in\{0, 1\}$ and $g^{(z)}_t(\dx| \xi)=\E[\mathbf{z} |\dx^t=\dx, \xi]$. We have that 
\begin{align*}
    b_t(\dx|\xi)&=\E[\dot{\alpha_t} \dx^0+\dot{\beta_t} \dx^1- {\gamma_t} \dot{\gamma_t}\nabla \log \rho_t|\dx^t=\dx, \xi]\\
    &=\dot{\alpha_t} g^{(0)}  + \dot{\beta_t} g^{(1)}  + \dot{\gamma_t} g^{(z)}\\
    &=\E[\dot{\alpha_t} \dx^0+\dot{\beta_t} \dx^1+\dot{\gamma_t} \mathbf{z} |\dx^t=\dx, \xi],
\end{align*} 
where the first equality follows by \eqref{FPK_Eqn} and the last one follows by taking derivative to \eqref{discrete_time} w.r.t. the index $t$.

We also observe that  $\nabla \log \rho_t=-\gamma_t^{-1} \E[ \mathbf{z} |\dx^t=\dx]$.

\qed
\end{proof}

\begin{theorem}\label{minimizers}
    The loss functions used for estimating the vector field follow that
    \begin{align*}
        L_i(\hat g^{(i)})=\int_0^1 \E[|\hat g^{(i)}|^2 - 2\dx^i \cdot \hat g^{(i)}]\dd t,
    \end{align*}
    where $i\in\{0, 1, z\}$, the expectation takes over the density for $(\dx^0, \dx^1)\sim \rho(\dx^0, \dx^1|\xi)$, $\xi\sim \eta(\xi)$, and $ \mathbf{z} \sim \mathcal{N}(\mathbf{0}, \mathbf{I})$.
\end{theorem}

\begin{proof}
    To show the loss is effective to estimate $g^{(0)}$, $g^{(1)}$, and $g^{(z)}$. It suffices to show
    \begin{align*}
        L_0(\hat g^{(0)})&=\int_0^1 \E[|\hat g^{(0)}|^2 - 2\dx^0 \cdot \hat g^{(0)}]\dd t,\\
        &=\int_0^1 \int_{\mathbb{R}^D}\bigg[|\hat g^{(0)}|^2 - 2\E[\dx^0|\dx^t=\dx, \xi] \cdot \hat g^{(0)}\bigg]\dd \dx \dd t,\\
        &=\int_0^1 \int_{\mathbb{R}^D}\bigg[|\hat g^{(0)}|^2 - 2 g^{(0)} \cdot \hat g^{(0)}\bigg]\dd \dx \dd t,
    \end{align*}
    where the last equality follows by definition. The unique minimizer is attainable by setting $\hat g^{(0)} = g^{(0)}$.

    The proof of $g^{(1)}$ and $g^{(z)}$ follows a similar fashion.

\end{proof}

\section{Experiment details} \label{appendix:experiment}

\subsection{Time series data}\label{app:ts_data}
The time series datasets include:
Solar \citep{lai2018modeling}, Exchange \citep{lai2018modeling}, 
Traffic\footnote{\url{https://archive.ics.uci.edu/ml/datasets/PEMS-SF}},  
and Wikipedia\footnote{\url{5https://github.com/mbohlkeschneider/gluon-ts/tree/mv_release/datasets}}.
We follow the preprocessing steps as in \cite{salinas2019high}.
Details of the datasets are listed in Table \ref{tab:ts_data}.

\begin{table}[H]
\caption{Properties of the datasets. }
\centering
\begin{tabular}{lcccc}
\toprule
Datasets & Dimension $D$ & Frequency & Total time points & Prediction length $P$ \\
\midrule
Exchange &  8 & Daily & 6,071 & 30 \\
Solar    &  137 & Hourly & 7,009 &  24 \\
Traffic   &  963 & Hourly & 4,001 &  24 \\
Wiki   &  2000 & Daily & 792 &  30 \\
\bottomrule
\end{tabular}
\label{tab:ts_data}
\end{table}

The probabilistic forecasting is evaluated by Continuous Ranked Probability Score (CRPS-sum) \citep{koochali2022random}, normalized root mean
square error via the median of the samples (NRMSE), and point-metrics normalized deviance (ND).
The metrics calculation is provided by \texttt{gluonts} package \citep{alexandrov2019gluonts} by calling module
\texttt{gluonts.evaluation.MultivariateEvaluator}.

\subsection{Models and hyperperameters}
Baseline models such as DDPM, SGM, FM, and SI all involve modeling field functions, where the inputs are the state vector (in the same space as the data samples), diffusion time, and condition embedding, and the output is the generated sample.
The field functions correspond to the ``noise prediction'' function in DDPM;
the score function in SGM;
the vector field in FM;
the velocity and score functions in SI.
To make a fair comparison between these models, we use the same neural networks for these models.

In the synthetic datasets experiments, we model the field functions with a 4-layer ResNet, each layer has 256 intermediate dimensions.
The batch size is 10,000 for all models and a model is trained with 20,000 iterations. The learning rate is $10^{-3}$.

In the time series forecasting experiments, the RNN for the history encoder has 1 layer and 128 latent dimensions;
The field function is modeled with a Unet-like structure \citep{ronneberger2015u} with 8 residual blocks, and each block has 64 dimensions.
To stabilize the training, we also use paired sampling for the stochastic interpolants introduced by \cite[Appendix C]{stochastic_interpolant}:
\begin{align*}
\mathbf{x}^s =& \alpha(s) \mathbf{x}^0 + \beta(s) \mathbf{x}^1 + \gamma(s) \mathbf{z} \\
{\mathbf{x}^{s}}' = & \alpha(s) \mathbf{x}^0 + \beta(s) \mathbf{x}^1 + \gamma(s) (-\mathbf{z}) \\
& s\in [0, 1], \mathbf{z}\sim \mathcal{N}(\mathbf{0}, \mathbf{I}).
\end{align*}
The baseline models are trained with 200 epochs and 64 batch sizes with a learning rate $10^{-3}$.
The SI model is trained with 100 epochs and 128 batch sizes with a learning rate $10^{-4}$. We find if the learning rate is too large, SI may not converge properly.

\subsection{Importance sampling }
The loss functions for training the velocity and score functions are

\begin{equation}
\begin{aligned}
\mathcal{L}_b=&\int_0^1 \mathbb{E}\Big[
\frac{1}{2}\| \hat{\mathbf{b}} (s,\mathbf{x}^s) \|^2 
- \big(\dot{\alpha}(s) \mathbf{x}^0 
+ \dot{\beta}(s) \mathbf{x}^1 
+ \dot{\gamma}(s)\mathbf{z} \big)^T \hat{\mathbf{b}}(s,\mathbf{x}^s)
\Big] ds,  \\
\mathcal{L}_s=& \int_0^1 \mathbb{E}\Big[
\frac{1}{2}\| \hat{\mathbf{s}} (s,\mathbf{x}^s) \|^2 
 + \gamma^{-1}\mathbf{z}^T \hat{\mathbf{s}}(s,\mathbf{x}^s)
\Big] ds.
\end{aligned}
\end{equation}
Both loss functions involve the integral over diffusion time $s \in [0, 1]$ in the form of 
\begin{equation}
\mathcal{L} = \int_0^1 l(s) ds
\approx \sum_i l(s_i), \quad s_i \sim \mathrm{Uniform}[0, 1].
\end{equation}

\begin{figure}[H]
\centering
\begin{subfigure}[b]{0.3\textwidth}
\centering
\includegraphics[width=\textwidth]{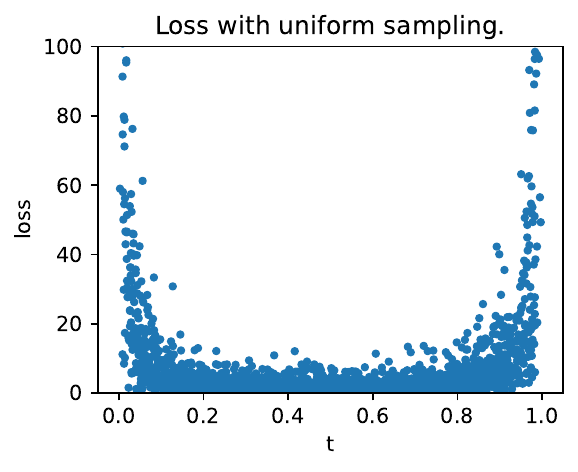}
\end{subfigure}
\begin{subfigure}[b]{0.32\textwidth}
\centering
\includegraphics[width=\textwidth]{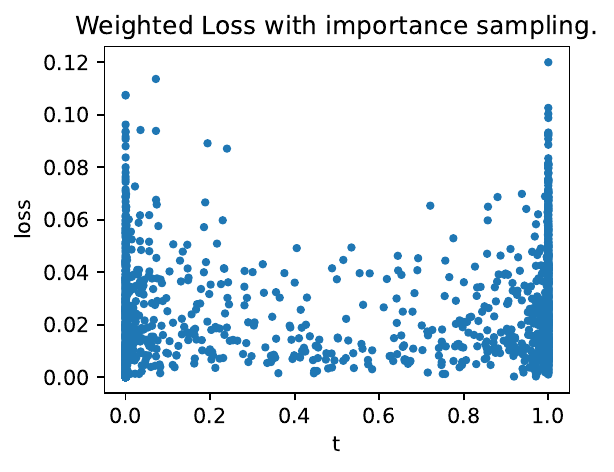}
\end{subfigure}
\caption{Comparison between uniform sampling and importance sampling. Each dot represents the loss of one sample with respect to the diffusion time. }
\label{fig:importance_sampling}
\end{figure}

However, the loss value $l(s)$ has a large variance, especially when $s$ is near $0$ or $1$. Figure \ref{fig:importance_sampling} shows an example of the distribution of $l(s)$ across multiple $s$. The large variance slows down the convergence of training.
To overcome this issue, we apply importance sampling, a similar technique used by \cite[Sec. 5.1 ]{song2021sde}, to stabilize the training.
Instead of drawing diffusion time from a uniform distribution, importance sampling considers,
\begin{equation}
\mathcal{L} = \int_0^1 l(s) ds
\approx \sum_i \frac{l(s_i)}{\tilde{q}(s_i)}, \quad s_i \sim \tilde{q}(s).
\end{equation}
Ideally, one wants to keep $l(s_i)/\tilde{q}(s_i)$ as constant as possible such that the variance of the estimation is minimum.
The loss value $l(s)$ is very large when $s$ is close to $0$ or $1$, and $l(s)$ is relatively flat in the middle, and the domain of $s$ is $[0, 1]$, so we choose Beta distribution $\mathrm{Beta}(s; 0.1, 0.1)$ as the proposal distribution $\tilde{q}$.
As shown in Figure \ref{fig:importance_sampling}, the values of $l(s_i)/\tilde{q}(s_i)$ are plotted against their $s$, which becomes more concentrated in a small range.

\newpage
\subsection{Additional forecasting results}

\begin{table*}[ht]
\centering
\begin{tabular}{lcccc}
 & Exchange rate & Solar & Traffic & Wiki \\
\midrule
DDPM & 0.011\scriptsize{$\pm$0.004} & 0.377\scriptsize{$\pm$0.061} & 0.064\scriptsize{$\pm$0.014} & 0.093\scriptsize{$\pm$0.023} \\
FM & 0.011\scriptsize{$\pm$0.001} & 0.445\scriptsize{$\pm$0.031} & 0.041\scriptsize{$\pm$0.002} & 80.624\scriptsize{$\pm$89.804} \\
SGM & 0.01\scriptsize{$\pm$0.002} & 0.388\scriptsize{$\pm$0.026} & 0.08\scriptsize{$\pm$0.053} & 0.122\scriptsize{$\pm$0.026} \\
SI & 0.008\scriptsize{$\pm$0.002} & 0.399\scriptsize{$\pm$0.065} & 0.089\scriptsize{$\pm$0.006} & 0.091\scriptsize{$\pm$0.011} \\
\end{tabular}
\caption{ND-sum. A smaller number indicates better performance.}
\label{tab:nd_sum} 
\end{table*}

\begin{table*}[ht]
\centering
\begin{tabular}{lcccc}
 & Exchange rate & Solar & Traffic & Wiki \\
\midrule
DDPM & 0.013\scriptsize{$\pm$0.005} & 0.72\scriptsize{$\pm$0.08} & 0.094\scriptsize{$\pm$0.029} & 0.123\scriptsize{$\pm$0.026} \\
FM & 0.014\scriptsize{$\pm$0.002} & 0.849\scriptsize{$\pm$0.072} & 0.059\scriptsize{$\pm$0.007} & 165.128\scriptsize{$\pm$147.682} \\
SGM & 0.019\scriptsize{$\pm$0.004} & 0.76\scriptsize{$\pm$0.066} & 0.109\scriptsize{$\pm$0.064} & 0.164\scriptsize{$\pm$0.03} \\
SI & 0.01\scriptsize{$\pm$0.003} & 0.722\scriptsize{$\pm$0.132} & 0.127\scriptsize{$\pm$0.003} & 0.117\scriptsize{$\pm$0.011} \\
\end{tabular}
\caption{NRMSE-sum. A smaller number indicates better performance.}
\label{tab:nrmse_sum}
\end{table*}

\subsection{Baseline Model using Unconditional SI}

Additionally, we introduced a new experiment to verify the necessity of conditional SI over unconditional SI. The unconditional SI diffuses from pure noise and does not utilize the prior distribution from the previous time point. In this case, the context for the prediction is provided exclusively by the RNN encoder. The new results are shown in the following tables. When compared with the conditional SI framework, the unconditional model shows slightly inferior performance.

\begin{table*}[ht]
\centering
\begin{tabular}{lccc}
 & Exchange rate & Solar & Traffic  \\
\midrule
SI & 0.007\scriptsize{$\pm$0.001} & 0.359\scriptsize{$\pm$0.06} & 0.083\scriptsize{$\pm$0.005} \\
Vanilla SI & 0.010\scriptsize{$\pm$0.001} & 0.383\scriptsize{$\pm$0.010} & 0.082\scriptsize{$\pm$0.006} 
\end{tabular}
\caption{CRPS-sum metric on multivariate probabilistic forecasting. A smaller number indicates better performance.}
\label{tab:CPRS_sum_vs_uncond_SI}
\end{table*}

\begin{table*}[!ht]
\centering
\begin{tabular}{lccc}
 & Exchange rate & Solar & Traffic  \\
\midrule
SI & 0.008\scriptsize{$\pm$0.002} & 0.399\scriptsize{$\pm$0.065} & 0.089\scriptsize{$\pm$0.006} \\
Vanilla SI & 0.010\scriptsize{$\pm$0.003} & 0.430\scriptsize{$\pm$0.113} & 0.093\scriptsize{$\pm$0.007} 
\end{tabular}
\caption{ND-sum. A smaller number indicates better performance.}
\label{tab:nd_sum_vs_uncond_SI}
\end{table*}

\begin{table*}[!ht]
\centering
\begin{tabular}{lccc}
 & Exchange rate & Solar & Traffic  \\
\midrule
SI & 0.010\scriptsize{$\pm$0.003} & 0.722\scriptsize{$\pm$0.132} & 0.127\scriptsize{$\pm$0.003} \\
Vanilla SI & 0.012\scriptsize{$\pm$0.003} & 0.815\scriptsize{$\pm$0.135} & 0.132\scriptsize{$\pm$0.015} 
\end{tabular}
\caption{NRMSE-sum. A smaller number indicates better performance.}
\label{tab:nrmse_sum_vs_uncond_SI}
\end{table*}

\end{document}